# A Randomized PDE Energy–driven Iterative Framework for Efficient and Stable PDE Solutions

Bing Yi[1*], Ran Zheng[1], Jinyang Fu[2,3*], Long Liu[4], Xiang Peng[5]

[1] School of Traffic and Transportation Engineering, Central South University, Changsha, China

[2] School of Civil Engineering, Central South University, Changsha, China

[3] National Engineering Research Center of High-speed Railway Construction Technology, Changsha, China

[4] School of Urban Construction and Transportation, Hefei University, Hefei, China

[5] College of Mechanical Engineering, Zhejiang University of Technology, Hangzhou, China

**Abstract:** Efficient and stable solution of partial differential equations (PDEs) is central to scientific and engineering applications, yet existing numerical solvers rely heavily on matrix-based discretizations, while learning-based methods require costly training and often suffer from limited generalization. In this work, we proposes a PDE energy–driven framework that solves PDEs through physically constrained diffusion iterations, without relying on classical matrix-based finite element assembly or data-driven neural network training. The proposed method evolves arbitrary random initial fields through PDE energy–driven implicit iterations combined with Gaussian smoothing, while strictly enforcing boundary conditions at each iteration. The proposed formulation is applied to representative one-dimensional Poisson, Heat, and viscous Burgers equations, covering both steady-state and transient problems. Numerical results demonstrate stable convergence to the unique physical solution from random initializations, with accurate resolution of sharp gradients and controlled Mean Squared Error (MSE) across a wide range of discretization parameters. Detailed comparisons with analytical solutions indicate that the framework achieves competitive accuracy and stability. Overall, the proposed framework provides a fast, flexible, and physically consistent alternative to traditional numerical solvers, offering a potential pathway for scalable PDE solutions in both research and engineering applications.

## 1. Introduction

Partial differential equations (PDEs) constitute the essential mathematical foundation for a vast range of scientific and engineering models, governing fundamental phenomena such as heat transfer, fluid dynamics, solid mechanics, and multiphysics interactions [1]. By bridging the gap between theoretical governing laws and practical implementation, accurate and efficient PDE solvers facilitate the critical transition from abstract physical principles to high-fidelity engineering analysis and design optimization [2]. As industries pivot toward full-life cycle Prognostics and Health Management (PHM) and Digital Twin paradigms, the functional role of these solvers has fundamentally evolved from static offline analysis to dynamic, real-time predictive engines [3]. In these complex environments, the challenge lies in assimilating multi-source environmental disturbances and transforming high-dimensional stochasticity into precise, deterministic trajectories for mission-critical forecasting [4]. Consequently, as problem scales grow and physical models become increasingly nonlinear and multiscale, the demand for robust, stable, and computationally efficient solution strategies has never been more pressing.

To meet these demands, a rich ecosystem of numerical and computational approaches has been developed over the past several decades. These approaches can be broadly categorized into classical discretization-based solvers, data-driven learning methods, physics-informed neural networks, and more recently, diffusion-based generative frameworks:

### a) Classical Numerical Methods

Classical numerical solvers for PDEs are dominated by finite difference methods (FDM), finite volume methods (FVM), and finite element methods (FEM). These approaches discretize the governing equations over spatial and temporal grids, resulting in large systems of algebraic equations whose solutions approximate the continuous PDE solution. Among them, FEM remains the gold standard for complex geometries and multiphysics problems due to its variational formulation and topological flexibility, forming the core of many commercial and open-source solvers in solid mechanics and fluid–structure interaction applications [5,6].

To alleviate the computational burden associated with large sparse matrix systems, multigrid and domain decomposition techniques have been extensively developed to accelerate convergence and enable scalable parallel implementations [7,8]. Spectral and pseudo-spectral methods offer exponential convergence for smooth solutions by employing global basis functions, though their applicability is limited by geometric complexity and boundary condition handling [9].

Despite their maturity and reliability, classical numerical methods face persistent challenges. Matrix assembly and inversion become prohibitively expensive for large-scale or high-dimensional problems, explicit schemes are constrained by stringent stability conditions such as the Courant–Friedrichs–Lewy (CFL) limit[10], and nonlinear PDEs often require iterative linearization, stabilization, and sophisticated preconditioning[11]. Moreover, accurately resolving localized sharp gradients, shocks, or boundary layers typically necessitates adaptive mesh refinement (AMR) or high-order schemes, substantially increasing implementation complexity and computational cost[12].

### b) Data-Driven and Operator Learning Approaches

Driven by the transformative advancements in deep learning, data-driven approaches have been proposed to approximate PDE solutions or operators directly from observational or simulated data, circumventing the need for explicit discretization. Initial surrogate models focused on training neural networks to map specific inputs—such as boundary conditions or physical parameters—to solution fields. However, these models were often constrained by the necessity for exhaustive labeled datasets and suffered from a marked inability to extrapolate beyond the distribution of the training regime.

To transcend these limitations, operator learning frameworks such as DeepONet and Fourier Neural Operators (FNO) have emerged as powerful tools for learning mappings between infinite-dimensional function spaces. DeepONet employs a branch–trunk architecture to approximate nonlinear operators and has demonstrated generalization across varying inputs and boundary conditions [13]. Concurrently, FNO leverages global spectral representations to learn solution operators through the integration of Fourier layers. This grid-independent approach has achieved remarkable success in modeling parametric PDE families and complex turbulent flows with near-instantaneous inference speeds [14,15].

Nevertheless, data-driven operator learning methods remain fundamentally "data-hungry", requiring vast repositories of high-fidelity training data that are predominantly synthesized via conventional numerical solvers. Their predictive performance often degrades when confronted with sharp gradients, extreme parameter regimes, or highly nonlinear transient dynamics. Most critically, these frameworks frequently lack inherent mechanisms to guarantee physical consistency, such as the exact enforcement of boundary conditions or the satisfaction of fundamental conservation laws. A comprehensive review of these mathematical foundations and their current limitations in learning mappings between function spaces is provided by Kovachki [16]. To address these challenges, the field is increasingly exploring the integration of prior physical knowledge into the learning process.

**c) Physics-Informed Neural Networks (PINNs)**

Physics-Informed Neural Networks (PINNs) represent a seminal conceptual shift by embedding the governing PDEs directly into the deep learning objective function. Rather than relying exclusively on large-scale labeled datasets, PINNs leverage automatic differentiation to minimize the residual of the PDE, alongside violations of initial and boundary conditions, within the network's loss function [17, 18]. This "physics-as-a-prior" paradigm enables mesh-free, and in some cases data-free, learning, proving exceptionally effective for solving high-dimensional inverse problems, parameter identification, and complex multiphysics coupling where observational data may be sparse or noisy.

To enhance the robustness and accuracy of the original formulation, numerous architectural extensions have been developed. These include Conservative PINNs (cPINNs), which explicitly enforce integral conservation laws across subdomains [19]; Adaptive PINNs, which utilize self-adaptive loss balancing and weight-adjustment strategies to navigate the often non-convex loss landscape [20]; and Domain Decomposition PINNs (XPINNs) designed to improve scalability and spatial resolution for multiscale phenomena [21].

Despite these algorithmic refinements, PINNs still encounter significant computational challenges. The training process is frequently plagued by slow convergence or optimization instabilities, high sensitivity to hyperparameter configurations, and a failure to resolve sharp gradients or capture long-term transient dynamics accurately. Furthermore, a standard PINN is typically a "point-wise" solver for a specific instance, any modification to the boundary conditions, domain geometry, or physical parameters necessitates a complete retraining of the network. This lack of inherent generalization remains a critical bottleneck for the rapid iterations required in real-time engineering design and Digital Twin maintenance.

**d) Diffusion Models and Physics-Augmented Generative Methods**

Diffusion-based generative frameworks, encompassing Denoising Diffusion Probabilistic Models (DDPMs) and score-based generative models, have redefined the state-of-the-art in modeling complex, high-dimensional data distributions. By conceptualizing data generation as a reversible, stochastic diffusion–denoising process, these models iteratively transform latent Gaussian noise into highly structured samples with exceptional stability and expressivity [22, 23].

Inspired by these breakthroughs, recent research has pivoted toward integrating physical priors into the generative pipeline. Emerging strategies primarily focus on posterior sampling for inverse problems via physics-informed guidance [24] and the high-fidelity synthesis of PDE solutions through conditional score-based frameworks [25]. However, these physics-augmented approaches typically necessitate the training of complex neural denoisers or score networks, thereby inheriting the intensive data dependencies, significant computational overhead, and potential training instabilities characteristic of deep generative architectures. Most critically, the strict enforcement of differential operators and boundary conditions is rarely guaranteed within the latent sampling process. Furthermore, the convergence of these stochastic frameworks toward a unique physical solution—as opposed to a broad probabilistic distribution—remains largely empirical and theoretically elusive [26].

In conclusion, while each paradigm offers distinct advantages, a critical gap persists between the rigorous reliability of classical numerical solvers and the expressive power of modern generative frameworks. Specifically, there is an ubiquitous need for a methodology that (i) inherits the robust, iterative refinement logic of diffusion-inspired processes, (ii) strictly enforces governing PDE operators and boundary conditions throughout the evolution, and (iii) bypasses both the prohibitive cost of large-scale matrix inversions and the intensive requirements of neural network training.

To address this gap, we propose a PDE energy-driven iterative framework designed for robust and real-time predictive tasks in Digital Twin ecosystems. By reformulating PDE solving as a physically constrained, stochastic-to-deterministic evolution process, our method initiates from randomized fields and iteratively refines the solution through updates guided by the variational PDE energy landscape. This architecture incorporates Gaussian regularization and exact boundary enforcement at each step, ensuring numerical stability and hardware-efficient execution. Unlike data-driven diffusion models, the heuristic denoising mechanism is entirely replaced by governing physical operators, enabling deterministic convergence to the unique physical solution without the need for high-fidelity training data or pre-trained parameters. Such a training-free and latency-aware approach provides a scalable pathway for integrating physical rigor into the high-frequency synchronization and robust decision-making cycles of modern digital twins.

The remainder of this paper is organized as follows. Section 2 presents the theoretical foundation of the Randomized PDE energy–driven framework. Section 3 details the numerical methodology for Poisson, Heat, and Burgers equations. Section 4 reports numerical results and comparative analyses with analytical solutions. Section 5 concludes with a discussion of limitations and future extensions toward higher-dimensional and multiphysics systems.

## 2. Theoretical Framework

### 2.1. Governing Equations and Problem Definition

Let $\Omega \in \mathbb{R}^d$ be a bounded spatial domain with boundary $\partial\Omega$. We consider a general class of time-dependent partial differential equations of the form:

$$\mathcal{L}[u](x,t) + N[u](x,t) = f(x,t) \qquad x \in \Omega, t \in (0,T] \qquad (1)$$

subject to Dirichlet boundary and initial conditions.

$$u(x,t)|_{\partial\Omega} = g(x,t), \qquad u(x,0) = u_0(x) \qquad (2)$$

Here, $\mathcal{L}$ denotes a linear differential operator (e.g., diffusion $\nu\Delta u$), $N$ represents nonlinear terms (e.g., convection $u \cdot \nabla u$), and $f(x,t)$ is a prescribed source term. This formulation encompasses elliptic, parabolic, and nonlinear transport–diffusion equations commonly encountered in scientific computing, including the Poisson, Heat, and viscous Burgers equations considered in this work [27-32]. The objective is to compute a stable and accurate numerical approximation of $u(x,t)$, while allowing initialization from random fields and avoiding both large-scale matrix assembly and data-driven training procedures.

### 2.2 Residual-Based Energy Functional

Instead of employing a direct discretization of the governing PDE into a rigid matrix system, we reformulate the problem as the minimization of a continuous residual-based energy functional:

$$E[u] = \frac{1}{2}\int_\Omega \|L[u] + N[u] - f\|^2 d\Omega \qquad (3)$$

where the exact PDE solution corresponds to a global minimizer of $E[u]$. This formulation is conceptually rooted in least-squares finite element methods and variational residual minimization, frameworks renowned for their inherent numerical robustness and coercivity properties [33].

The variation of $E$ with respect to $u$ yields the functional gradient, which defines the descent

direction for our iterative process:

$$\frac{\delta E}{\delta u} = L^*(L[u] + N[u] - f) + \left(\frac{\partial N[u]}{\partial u}\right)^* (L[u] + N[u] - f) \tag{4}$$

where $L^*$ denotesthe adjoint of the linear operator, and $\left(\frac{\partial N[u]}{\partial u}\right)^*$ represents the adjoint of the Fréchet derivative of the nonlinear term. For purely linear operators where ($N = 0$), simplifies to:

$$\frac{\delta E}{\delta u} = L^*(L[u] - f) \tag{5}$$

This linear reduction ensures a symmetric and coercive operator for the optimal system, effectively transforming even non-self-adjoint problems into a symmetric, positive-definite (SPD) descent landscape. Such a structural advantage provides the mathematical guarantee for stable convergence from arbitrary, randomized initializations.

2.3 Randomized Initialization and Deterministic Convergence

The solution is initialized from a randomized field:

$$u_0(x) = u_{\text{rand}}(x), u_{\text{rand}} \sim \mathcal{N}(0, \sigma^2) \tag{6}$$

where $\sigma$ controls the initial noise amplitude. Unlike stochastic solvers or learning-based diffusion models, this randomness does not affect the final solution. Instead, it serves as a generic initial condition from which the deterministic PDE-driven evolution converges toward the unique physical solution. Such insensitivity to initial conditions is a characteristic property of dissipative PDE systems with unique steady-state attractors, as established in the theory of infinite-dimensional dynamical systems and nonlinear evolution equations [34-36].

As the iterative process progresses, the transition from a stochastic initial field to a deterministic physical solution is mathematically described by a Langevin-type Stochastic Differential Equation (SDE) [23, 37]:

$$du = -\frac{\delta E}{\delta u} dt + \sqrt{2\varepsilon}\, \xi, \quad \xi \sim N(0, I) \tag{7}$$

where $\varepsilon$ represents the noise intensity (analogous to temperature in statistical mechanics). This equation characterizes the trajectory in function space as a competition between the gradient descent toward the energy minimum and a vanishing thermal fluctuation. The long-term behavior of this system is governed by the Fokker–Planck equation, which ensures that the probability density of the evolving solution admits a stationary Gibbs measure [38, 39]:

$$p_\infty(u) \propto exp\left(-\frac{E(u)}{\varepsilon}\right) \tag{8}$$

This formulation provides a rigorous bridge between modern score-based generative models and the variational geometry of dissipative evolution. In our framework, the iterative update acts as an annealing process: as the noise level $\varepsilon \to 0$, the distribution $p_\infty(u)$ collapses to a Dirac delta mass concentrated precisely on the exact solution of the PDE. This ensures that regardless of the initial noise, the framework achieves deterministic convergence to the unique physical state consistent with the governing equations.

2.4 Implicit Gradient Flow in Energy Space

We define an artificial evolution equation in pseudo-time $\tau$ as the gradient flow in the energy space. To ensure numerical stability while maintaining the statistical properties of the Langevin dynamics, we adopt a semi-implicit discretization:

$$\frac{u^{n+1}-u^n}{\Delta\tau} = -\frac{\delta E}{\delta u}\Big|_{u^{n+\theta}} + \sqrt{\frac{2\varepsilon}{\Delta\tau}}\,\xi^n, \quad \xi^n \sim N(0,I) \tag{9}$$

Linearizing the functional gradient yields the linearized update system:

$$(I - \Delta\tau J)u^{n+1} = u^n + \Delta\tau b^n + \sqrt{2\varepsilon\Delta\tau}\,\xi^n \tag{10}$$

where $J$ represents the Jacobian matrix of the energy functional E(or a suitable linear approximation), and $b^n$ accounts for nonlinear terms and source contributions. For linear elliptic and parabolic PDEs, this scheme is unconditionally stable, permitting large pseudo-time steps that bypass the restrictive CFL constraints typical of explicit schemes[40].

### 2.5 Gaussian Regularization as Scale Control

To suppress high-frequency oscillations introduced by random initialization or nonlinear advection, a smoothness regularization operator is applied at each iteration:

$$\tilde{u}^{n+1} = S[u^{n+1}] = G_\sigma * u^{n+1} \tag{11}$$

where $G_\sigma$ is a Gaussian kernel. This operation acts as a selective spectral filter, analogous to artificial viscosity commonly used in numerical stabilization [41,42]. Importantly, the regularization is applied after the PDE-consistent update and therefore does not alter the fixed-point solution of the governing equations.

### 2.6 Exact Boundary Condition Enforcement

Unlike penalty-based methods or weak constraints common in Physics-Informed Neural Networks (PINNs), Dirichlet boundary conditions are enforced exactly and explicitly at each iteration:

$$u^{n+1}(\mathbf{x})|_{\partial\Omega} = g(\mathbf{x}, t^{n+1}) \tag{12}$$

This step ensures that the solution trajectory remains within the physically admissible manifold of the function space throughout the entire evolution [43]. This rigorous enforcement distinguishes our method from learning-based solvers, which often struggle with "leaky" boundaries or require delicate hyperparameter tuning for penalty terms.

## 3. Numerical Implementation

This section presents the concrete numerical realization of the proposed PDE energy–driven iterative framework for three representative equations: the Poisson equation, the heat equation, and the viscous Burgers equation. For each case, we explicitly derive the discrete residual, the energy functional, and the corresponding iterative update rule consistent with the Langevin-type gradient flow established in Section 2.

### 3.1 Poisson Equation

#### 3.1.1 Governing equation and discretization

We consider the Poisson equation on a square domain $\Omega = [0,1]^2$:

$$-\Delta u = f \quad \text{in } \Omega, \qquad u = 0 \quad \text{on } \partial\Omega \tag{13}$$

The domain is discretized using a uniform grid with spacing $h$. A standard second-order central difference scheme is employed to approximate the Laplacian:

$$\Delta_p u_{i,j} \approx \frac{u_{i+1,j} + u_{i-1,j} + u_{i,j+1} + u_{i,j-1} - 4u_{i,j}}{h^2} \tag{14}$$

#### 3.1.2 Residual and energy functional

Consistent with the Least-Squares formulation, we define the discrete residual as:

$$r(u) = -\Delta_p u - f \tag{15}$$

The corresponding energy functional E(u) represents the $L_2$-norm of this residual:

$$E_p(u) = \frac{1}{2}\|r(u)\|_2^2 = \frac{1}{2}\|-\Delta_p u - f\|_2^2 \tag{16}$$

The functional gradient is derived as:

$$\nabla E_p(u) = \Delta_p^*(\Delta_p u + f) \tag{17}$$

where $\Delta_p^*$ denotes the adjoint operator of the discrete Laplacian. Under homogeneous Dirichlet boundary conditions, $\Delta_p$ is self-adjoint ($\Delta_p^* = \Delta_p$), and the composition $\Delta_p^*\Delta_p$ results in a symmetric positive definite (SPD) discrete biharmonic operator. This structure ensures that the energy landscape is strictly convex, providing a robust gradient descent direction.

### 3.1.3 Stochastic Implicit Update Rule

In accordance with the Langevin-driven gradient flow established in Section 2.4, the update rule incorporates both the implicit gradient descent and a stochastic forcing term to ensure global exploration of the function space:

$$\frac{u^{n+1}-u^n}{\Delta\tau} = -\Delta_p^*(\Delta_p u^{n+1} + f) + \sqrt{\frac{2\varepsilon}{\Delta\tau}}\,\xi^n, \quad \xi^n \sim N(0, I) \tag{18}$$

Rearranging the terms leads to the following linearized optimality system for each iteration:

$$(I + \Delta\tau\Delta_p^*\Delta_p)u^{n+1} = u^n - \Delta\tau\Delta_p^* f + \sqrt{2\varepsilon\Delta\tau}\,\xi^n \tag{19}$$

This formulation transforms the second-order Poisson problem into a fourth-order elliptic system. The SPD nature of the operator $I + \Delta\tau\Delta_p^*\Delta_p$ guarantees numerical stability even for large pseudo-time steps $\Delta\tau$. Following each implicit update, Gaussian smoothing $S[u]$ and exact boundary enforcement are applied to suppress high-frequency noise and maintain physical admissibility.

## 3.2 Heat Equation
### 3.2.1 Governing equation

To demonstrate the framework's capability in handling time-dependent processes, we consider the classical transient diffusion equation:

$$\frac{\partial u}{\partial t} - \nabla \cdot (\alpha\nabla u) = f(x,t), x \in \Omega, t \in [0, T] \tag{20}$$

where $u(x,t)$ represents the temperature field, $\alpha$ denotes the thermal diffusivity, and $f(x,t)$ is an external source term. Traditional solvers march through discrete time steps, which can lead to cumulative temporal errors. Our framework instead adopts a holistic spatio-temporal perspective.

### 3.2.2 Space-Time Stationary Transformation

We reframe the temporal dimension $t$ as a spatial-like coordinate $y \in [0, Y]$, thereby transforming the transient evolution into a stationary boundary value problem on a unified space-time domain $\mathcal{D} = \Omega \times [0, Y]$:

$$u_y - \alpha u_{xx} = f(x, y) \tag{21}$$

In this formulation, the physical constraints are partitioned into the governing equation within $\mathcal{D}$ and specific boundary requirements on $\partial\mathcal{D}$. The initial condition is treated as a Dirichlet constraint at the bottom edge ($y = 0$):

$$u(x, 0) = h(x), x \in [0,1] \tag{22}$$

while the spatial boundary conditions are defined along the lateral edges ($x \in \partial\Omega$):

$$u(x, y)|_{x \in \partial\Omega} = g(x, y) \tag{23}$$

### 3.2.3 Operator construction and energy functional

To implement the spatio-temporal operator $L_h = \partial_y - \alpha\partial_{xx}$ on a discretized grid, we define the system matrix A using the Kronecker product ($\otimes$):

$$A = (D_y \otimes I_x) - (I_y \otimes \alpha\Delta_h) \tag{24}$$

where $D_y$ and $\Delta_h$ represent the discrete first-order temporal and second-order spatial derivatives, respectively. The governing physics within the domain are enforced through a residual-based energy functional:

$$E_h(u) = \frac{1}{2} \| Au - f \|^2 \tag{25}$$

The gradient of this functional is given by:

$$\nabla E_h(u) = A^*(Au - f) \tag{26}$$

where $A^*$ is the adjoint of the discrete spatio-temporal operator.

### 3.2.4 Stochastic Dynamics and Implicit Iteration

Consistent with the theoretical framework in Section 2, the spatio-temporal field evolves toward the physical solution via a Langevin-type SDE:

$$du = -\nabla E_h(u)dt + \sqrt{2\varepsilon}\, \xi, \quad \xi \sim N(0, I) \tag{27}$$

Discretizing this flow implicitly yields the update rule:

$$(I + \Delta\tau A^*A)u^{n+1} = u^n + \Delta\tau A^* f + \sqrt{2\varepsilon\Delta\tau}\, \xi^n \tag{28}$$

The composition $A^*A$ effectively transforms the originally non-symmetric spatio-temporal operator—whose asymmetry arises from the first-order temporal derivative $u_y$—into a symmetric positive-definite (SPD) energy operator. This transformation ensures absolute stability and provides a robust curvature for the gradient descent. Within this framework, even as the deterministic residual approaches zero, the stochastic term $\xi^n$ maintains persistent statistical fluctuations, enabling the system to exhaustively explore the function space until the probability distribution collapses into a Dirac delta mass at the exact physical solution. To conclude each iteration, Gaussian smoothing and exact boundary enforcement are applied as a combined stabilization step, suppressing high-frequency spectral noise and ensuring that the solution trajectory strictly adheres to the prescribed Dirichlet constraints.

## 3.3 Viscous Burgers Equation
### 3.3.1 Governing equation and discretization

Finally, we consider the nonlinear Burgers' equation, a fundamental model for capturing the interaction between nonlinear advection and viscous diffusion:

$$\frac{\partial u}{\partial t} + u\frac{\partial u}{\partial x} - \nu\frac{\partial^2 u}{\partial x^2} = f, x \in \Omega, t \in [0, T] \tag{29}$$

where $u\frac{\partial u}{\partial x}$ is the nonlinear advection term, which tends to generate shock waves, $\nu\frac{\partial^2 u}{\partial x^2}$ is the linear diffusion term, which acts to smooth out the solution. This equation is characterized by the formation of sharp gradients and shocks, posing a significant challenge for traditional numerical solvers due to strong nonlinearity.

### 3.3.2 Space-Time Stationary Transformation

Consistent with our holistic approach, we transform the transient Burgers' equation into a stationary boundary value problem by mapping the temporal dimension $t$ to a spatial-like coordinate

$$u_y + uu_x - \nu u_{xx} = f \tag{30}$$

The physical constraints are defined by the initial state $u(x, 0) = h(x)$ and the spatial boundary conditions:

$$u(x, y)|_{x \in \partial\Omega} = g(x, y) \tag{31}$$

### 3.3.3 Residual and energy functional

We define the nonlinear spatio-temporal operator $\mathcal{F}(\mathbf{u})$ as:
$$\mathcal{F}(u) = (D_y \otimes I_x)u + u \odot [(I_y \otimes D_x)u] - \nu(I_y \otimes D_{xx})u \tag{32}$$

The discrete residual is $r(u) = \mathcal{F}(u) - f$, and the iterative process is devoted to minimizing the residual-based energy functional:
$$E_b(u) = \frac{1}{2} \| \mathcal{F}(u) - f \|^2 \tag{33}$$

Unlike linear cases, the gradient of $E_b(u)$ requires the Jacobian matrix $J(u)$ of the nonlinear operator. For the advection term $u\frac{\partial u}{\partial x}$, the derivative is given by:
$$\frac{\partial}{\partial u}(u \odot [(I_y \otimes D_x)u]) = diag((I_y \otimes D_x)u) + diag(u)(I_y \otimes D_x) \tag{34}$$

Consequently, the full Jacobian matrix for the Viscous Burgers system is expressed as:
$$J(u) = (D_y \otimes I_x) + diag((I_y \otimes D_x)u) + diag(u)(I_y \otimes D_x) - \nu(I_y \otimes D_{xx}) \tag{35}$$

This Jacobian construction captures the dual role of the velocity field: it acts both as the advected quantity and as the local advection velocity, providing the necessary sensitivity information for the implicit solver.

### 3.3.4 Stochastic Levenberg-Marquardt Update

To address the strong nonlinearity and ensure robust convergence toward the global energy minimum, we adopt a Stochastic Levenberg-Marquardt (LM) update rule. By linearizing the gradient flow and incorporating the Langevin-type stochastic forcing, the implicit update is formulated as:
$$(I + \Delta\tau J^*J)u^{n+1} = u^n + \Delta\tau J^*f + \sqrt{2\varepsilon\Delta\tau}\,\xi^n \tag{36}$$

In this formulation, the composition $J^*J$ functions as a symmetric positive-definite (SPD) energy operator, which effectively self-regularizes the sharp gradients and potential shocks inherent in Burgers' solutions. The left-hand side provides a damped, biharmonic-like regularization that stabilizes the inversion of the non-convex energy landscape.

The right-hand side integrates the previous state with the source term projected via the adjoint $J^*$, ensuring that each step is driven by the linearized physical constraints. Even as the system approaches physical equilibrium, the stochastic term $\xi^n$ maintains persistent fluctuations, preventing the iteration from stagnating in local suboptimal states or capturing spurious numerical artifacts. Following each update, Gaussian smoothing and exact boundary enforcement are applied to the updated field $u^{n+1}$ to maintain spectral stability and ensure the solution trajectory captures the delicate balance between nonlinear steepening and viscous dissipation.

## 4. Numerical Experiments and Discussion

This section presents a comprehensive evaluation of the proposed PDE energy-driven iterative framework. The evaluation is conducted across three mathematically distinct categories: the Poisson equation (Elliptic), the Heat equation (Parabolic), and the Viscous Burgers equation (Nonlinear Hyperbolic).The primary objective of these experiments is to demonstrate the framework's unique ability to achieve global convergence toward the exact solution, starting from high-entropy, stochastic initializations.

### 4.1 Poisson Equation: Global Convergence from Random Initialization
### 4.1.1 Convergence process and comparison with analytical solution

The Poisson equation serves as the primary benchmark for verifying the efficacy of the proposed steady-state reconstruction. Distinguishing itself from classical iterative solvers that mandate a near-equilibrium initial guess, the proposed framework demonstrates a robust path-independent convergence property.

Fig. 1 illustrates the iterative evolution starting from a completely random white-noise field. The framework effectively functions as a "physical sieve": during the incipient stage (e.g., within 10 iterations), it rapidly identifies the global topological structure of the potential field. As iterations progress, the embedded PDE-energy guidance systematically refines localized gradients. This process causes the numerical field to progressively "crystallize" into the analytical solution, with the global residual decaying monotonically toward a stable steady state by the 200th iteration..

The final steady-state result is presented in Fig. 2. A quantitative cross-sectional comparison at the domain midpoint (x = 0.5, as shown in Fig. 2b) demonstrates an exceptional level of alignment.The framework captures high-curvature peaks with high fidelity and zero spurious oscillations, achieving a relative $L^2$ error of 1.78%. These results underscore that the randomized inference strategy reliably identifies the unique solution manifold dictated by the governing operator without compromising numerical precision.

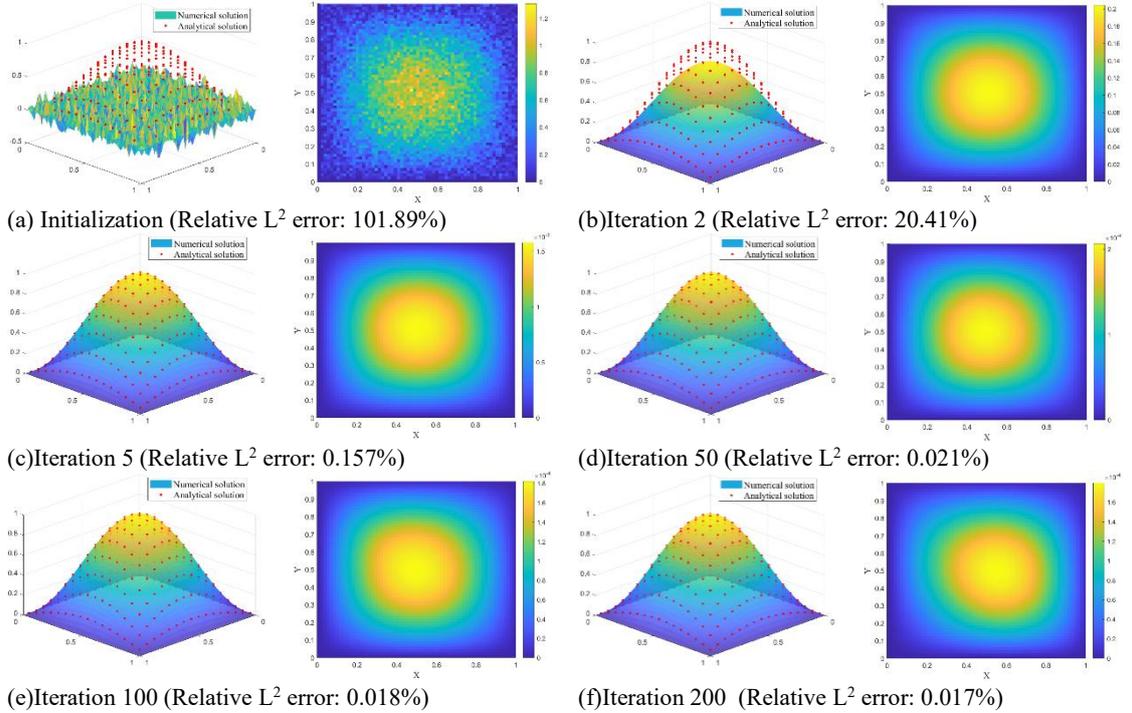

(a) Initialization (Relative $L^2$ error: 101.89%)  (b)Iteration 2 (Relative $L^2$ error: 20.41%)

(c)Iteration 5 (Relative $L^2$ error: 0.157%)  (d)Iteration 50 (Relative $L^2$ error: 0.021%)

(e)Iteration 100 (Relative $L^2$ error: 0.018%)  (f)Iteration 200 (Relative $L^2$ error: 0.017%)

Fig. 1 Iterative convergence snapshots of the Poisson potential field evolving from high-entropy stochastic initializations (left: Comparison of the reconstructed field and the analytical solution, right:Spatial distribution of absolute error)

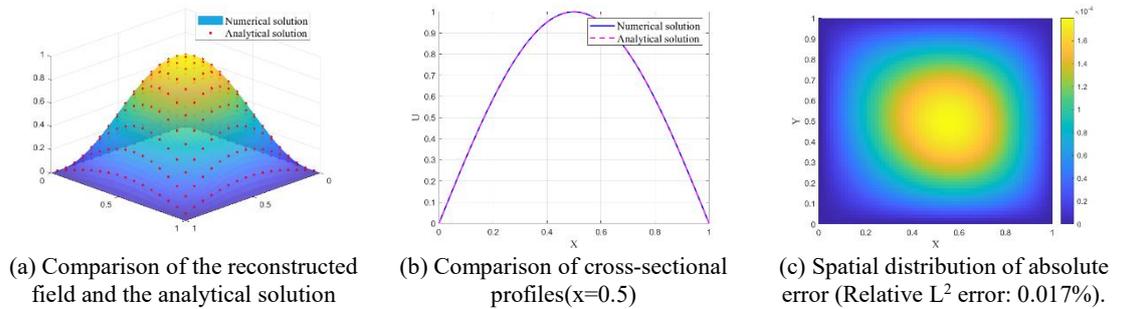

(a) Comparison of the reconstructed field and the analytical solution  (b) Comparison of cross-sectional profiles(x=0.5)  (c) Spatial distribution of absolute error (Relative $L^2$ error: 0.017%).

Fig.2 Steady-state reconstruction and accuracy verification for the Poisson equation

### 4.1.2 Role of Gaussian smoothing and boundary enforcement

To dissect the contribution of each algorithmic component to the global convergence, an ablation study was conducted under four distinct configurations. Fig. 3 illustrates the corresponding convergence trajectories (comprising PDE residuals and relative $L^2$ errors), cross-sectional profiles, comparison of the reconstructed field and the analytical solution and the spatial distributions of absolute error.

The convergence curves reveal that the PDE residuals across all four scenarios remain within a similar order of magnitude. This is primarily because the stochastic noise coefficient ($\varepsilon$) in the SDE driven prior is kept at a moderate level in these trials. Consequently, the impact of Gaussian smoothing on the final precision appears marginal in this specific parameter regime. However, it must be emphasized that Gaussian smoothing acts as a spectral filter that is vital for ensuring optimization stability in the presence of high-intensity initial noise.The experimental results suggest a necessary balance between stability and precision: while smoothing ensures smooth, monotonic convergence (Fig.3c), its absence can occasionally lead to even lower residuals (0.003% in Fig.3d) at the cost of potential instability under more chaotic initializations.

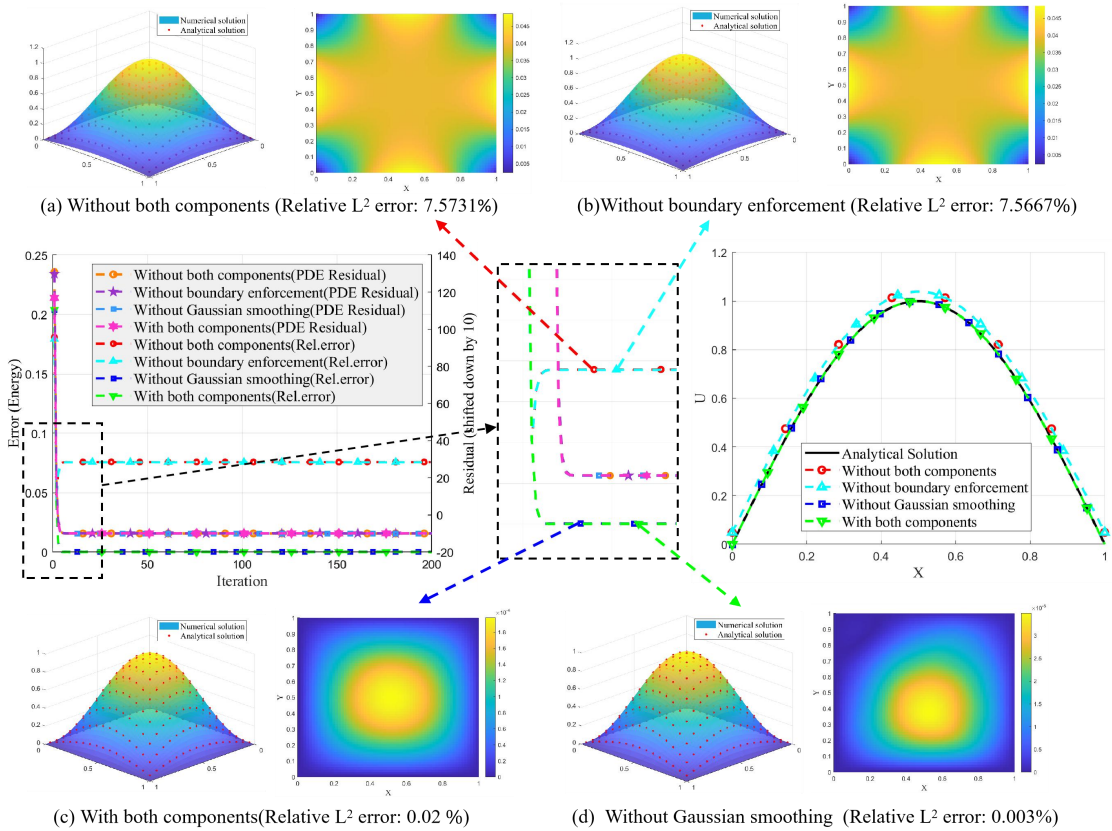

(a) Without both components (Relative $L^2$ error: 7.5731%)  (b) Without boundary enforcement (Relative $L^2$ error: 7.5667%)

(c) With both components (Relative $L^2$ error: 0.02 %)  (d) Without Gaussian smoothing (Relative $L^2$ error: 0.003%)

Fig. 3 Ablation study of Gaussian smoothing and boundary enforcement on convergence stability and solution uniqueness.

In contrast to the smoothing component, explicit boundary enforcement is found to be the decisive factor for physical exactness. As shown in the convergence history for the cases without boundary enforcement (Fig. 3a, b), although the PDE residual decreases rapidly, the relative $L^2$ error exhibits a paradoxical rebound behavior. After a sharp initial drop, the error begins to rise significantly around the 10th iteration, eventually stabilizing at a high plateau of approximately 7.5%. This phenomenon occurs for that the solver identifies a general solution that minimizes the PDE residual energy but fails to satisfy the uniqueness principle without Dirichlet boundary constraints. As evidenced by the cross-sectional profiles in Fig. 3, the resulting solution field exhibits a constant offset from the analytical benchmark. The solver effectively drifts on the solution manifold, satisfying the governing operator but violating the specific boundary values.

In contrast, when boundary manifold projection is enabled (Fig 3(c) and 3(d)), the relative $L^2$ error decays in lockstep with the PDE residual, rapidly converging to the unique exact solution. For the full framework, a steady-state relative error of 0.02% is achieved. This validates that while the PDE-energy guidance provides the driving force toward the solution manifold, the boundary constraints act as the "anchor" that isolates the particular solution from the infinite set of potential candidates. These results demonstrate that Gaussian smoothing serves as a critical stabilization mechanism for the iterative descent, whereas explicit boundary enforcement is indispensable for ensuring the physical consistency and mathematical uniqueness of the reconstructed field.

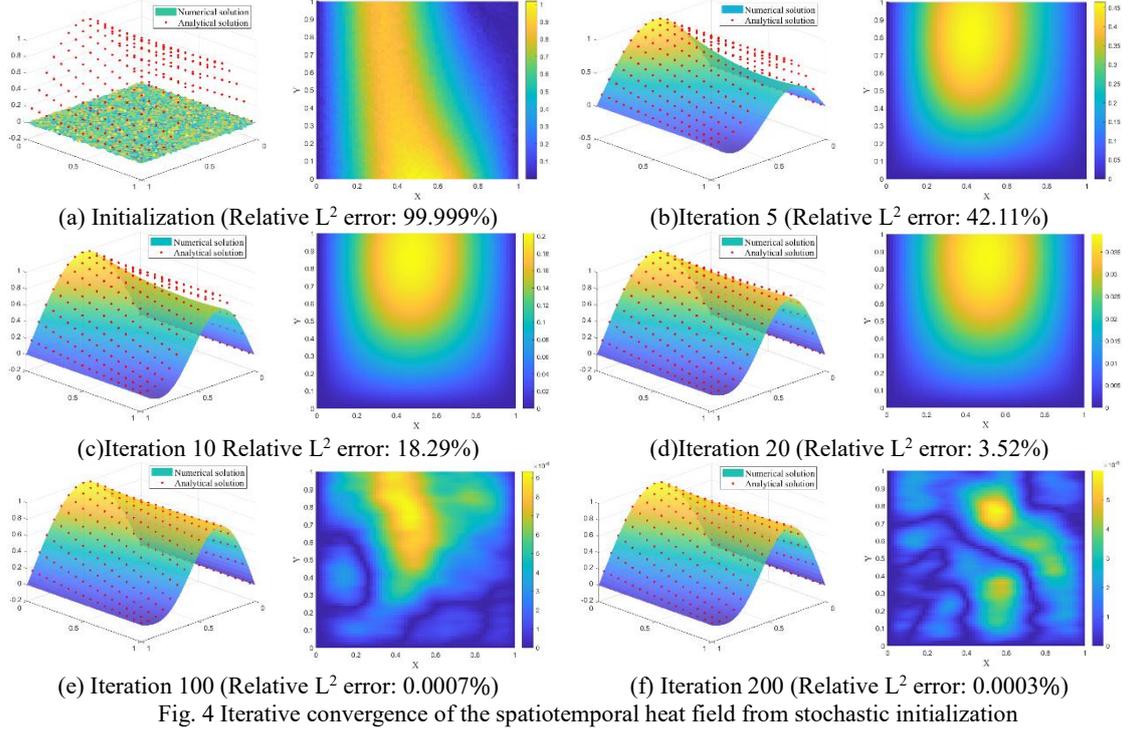

(a) Initialization (Relative $L^2$ error: 99.999%)      (b) Iteration 5 (Relative $L^2$ error: 42.11%)

(c) Iteration 10 Relative $L^2$ error: 18.29%)      (d) Iteration 20 (Relative $L^2$ error: 3.52%)

(e) Iteration 100 (Relative $L^2$ error: 0.0007%)      (f) Iteration 200 (Relative $L^2$ error: 0.0003%)

Fig. 4 Iterative convergence of the spatiotemporal heat field from stochastic initialization

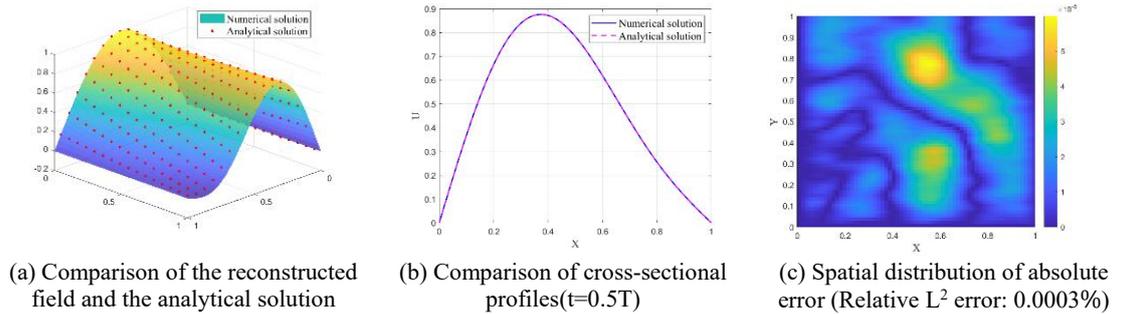

(a) Comparison of the reconstructed field and the analytical solution      (b) Comparison of cross-sectional profiles(t=0.5T)      (c) Spatial distribution of absolute error (Relative $L^2$ error: 0.0003%)

Fig.5 Steady-state thermal reconstruction and error distribution analysis

### 4.2 Heat Equation: Energy-Consistent Implicit Time Integration
#### 4.2.1 Convergence process and comparison with analytical solution

The transient heat equation, as defined in Section 3.2, is employed to evaluate the framework's performance on parabolic systems. A distinctive feature of the proposed method is the reformulating of the transient evolution into a static spatiotemporal manifold optimization. By treating time as a coordinate dimension, the solver bypasses the cumulative errors and stability constraints (e.g., CFL conditions) inherent in traditional sequential time-marching schemes.

Fig. 4 illustrates the iterative evolution of the spatiotemporal solution field initiated from a stochastic state. Consistent with the observations in the Section 4.1, the numerical field exhibits a rapid crystallization process. In the incipient iterations, the diffusion prior establishes the macroscopic thermal gradients across the spatiotemporal domain, while the subsequent energy-driven iterations refine the local temporal-spatial coupling. The global residual decays monotonically, demonstrating that the solver accurately captures the heat diffusion logic within the unified energy landscape.

The final reconstruction and accuracy metrics are presented in Fig. 5. A quantitative cross-sectional comparison at t = 0.5T and the spatial error distribution confirm that the framework achieves high fidelity in capturing the thermal decay profiles. Despite the increased complexity of the spatiotemporal manifold, the relative $L^2$ error is maintained at 0.0003%. This result proves that the proposed method can rigorously solve transient problems as static optimization tasks without sacrificing physical accuracy or temporal consistency.

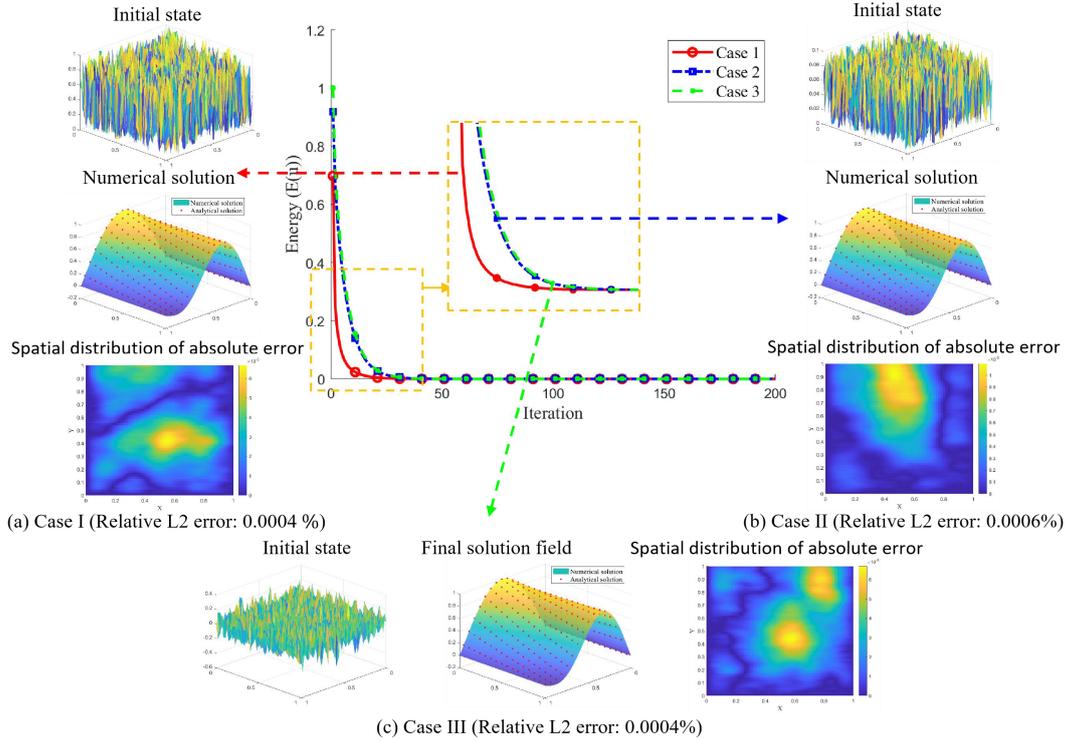

Fig.6 Verification of the global attractor property: Convergence trajectories from multiple random initializations.

### 4.2.2 Robustness with respect to arbitrary initial conditions

To further investigate the reliability of the energy-driven iteration, we assess its robustness against arbitrary initial conditions. Multiple simulations were conducted using stochastic initial fields with varying random seeds and magnitudes (Cases I, II, and III), as depicted in Fig. 6.

Numerical results demonstrate that all realizations consistently converge to the unique physical solution, irrespective of the chaotic nature or the magnitude of the initial noise. While individual convergence trajectories diverge—as the solver navigates disparate regions of the high-dimensional energy landscape—the final reconstructed fields achieve exceptional consistency. Specifically, the relative $L^2$ errors are strictly maintained within a narrow range of 0.0004% to 0.0006% compared to the analytical benchmark.

This behavior confirms the global attractor property of the proposed framework: for coercive parabolic operators, the residual-based energy landscape possesses a unique global minimizer. The iterative process reliably attracts any arbitrary initial state into this physical basin, ensuring that the final solution is entirely independent of the starting guess. Such path-independent convergence underscores the framework's potential for robust engineering applications, particularly in complex transient evolutionary problems where a reliable initial guess of the solution field is typically unavailable.

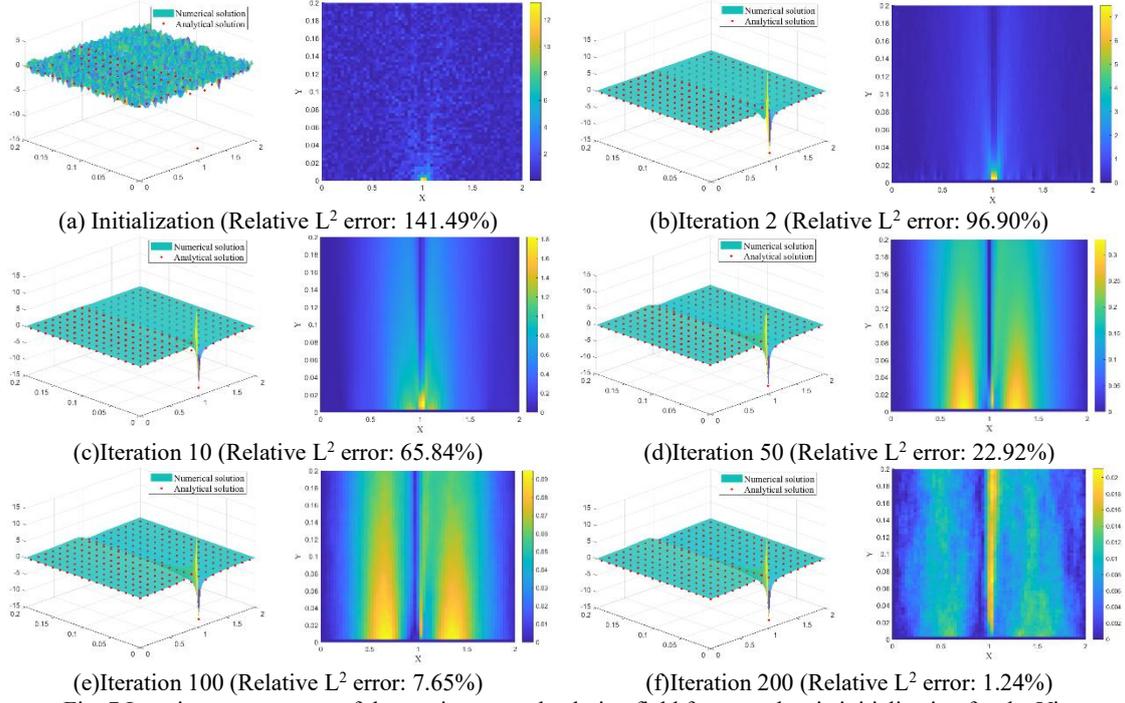

(a) Initialization (Relative $L^2$ error: 141.49%)　　(b)Iteration 2 (Relative $L^2$ error: 96.90%)

(c)Iteration 10 (Relative $L^2$ error: 65.84%)　　(d)Iteration 50 (Relative $L^2$ error: 22.92%)

(e)Iteration 100 (Relative $L^2$ error: 7.65%)　　(f)Iteration 200 (Relative $L^2$ error: 1.24%)

Fig. 7 Iterative convergence of the spatiotemporal solution field from stochastic initialization for the Viscous Burgers Equation

## 4.3 Viscous Burgers Equation: Nonlinearity and Robustness Analysis
### 4.3.1 Convergence behavior in nonlinear regimes

Finally, we evaluate the framework on the viscous Burgers equation, a canonical benchmark characterized by the intricate coupling of strong nonlinear convection and diffusion. The intrinsic difficulty in resolving this equation arises from the emergence of steep gradients and shock-like structures, which frequently trigger spurious numerical oscillations or catastrophic instabilities in traditional solvers. This experiment, therefore, serves a dual purpose: it rigorously validates the framework's capacity to navigate highly non-convex nonlinear operators while simultaneously extending the transient-to-static spatiotemporal manifold paradigm to the realm of non-equilibrium physics.

Fig. 7 captures the iterative evolution of the velocity field from a state of high-entropy stochastic perturbation. Despite the inherent challenges of nonlinear steepening, the framework demonstrates robust global convergence. It is noteworthy that by treating the transient Burgers evolution as a unified (d+1) dimensional steady-state problem, the solver effectively crystallizes the entire history of the shock formation from chaotic noise. The global residual decays monotonically, proving that the energy-driven guidance can navigate the complex, non-convex landscapes typical of nonlinear PDEs to identify the true physical manifold.

The definitive reconstruction and accuracy validation are presented in Fig. 8. Even in this challenging nonlinear regime, the framework achieves a relative $L^2$ error of 1.24%. While this is slightly higher than in the linear cases, it remains remarkably low given the presence of sharp gradients. The cross-sectional slice at t = 0.5 T in Fig. 8b confirms that the solver precisely identifies the shock front's position and slope without any manual hyperparameter tuning. This inherent shock-capturing capability, achieved starting from a purely random initialization, highlights the framework's potential for solving complex fluid dynamics where traditional time-marching schemes often suffer from numerical dissipation or instability.

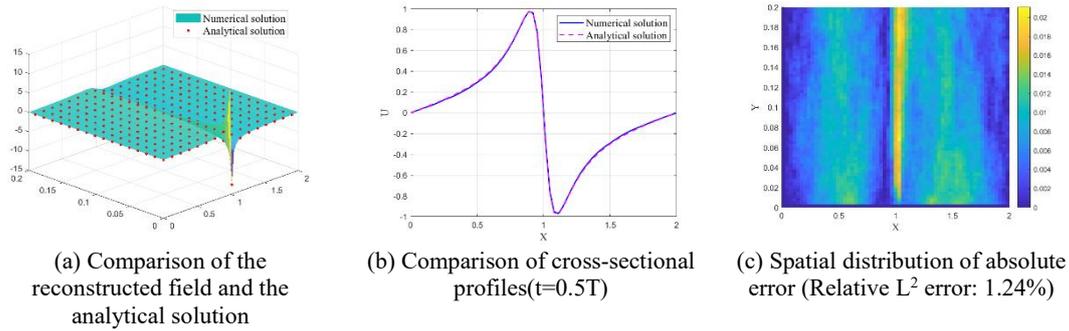

(a) Comparison of the reconstructed field and the analytical solution

(b) Comparison of cross-sectional profiles(t=0.5T)

(c) Spatial distribution of absolute error (Relative $L^2$ error: 1.24%)

Fig.8 Spatiotemporal reconstruction and accuracy verification for the viscous Burgers equation

### 4.3.2 Sensitivity Analysis and Statistical Reliability

**1) Impact of the Viscosity Coefficient $\nu$**

The diffusion coefficient $\nu$ is the critical parameter governing the balance between nonlinear convection and diffusion in the Burgers equation. We investigated the framework's performance across three distinct physical regimes: convection-dominated ($\nu = 0.01$, characterized by a sharp shock front), transition ($\nu = 0.05$), and diffusion-dominated ($\nu = 0.1$, characterized by a smooth profile).

As illustrated in Fig.9, the solver demonstrates exceptional adaptability to varying physical characteristics through a comprehensive comparison of reconstructed fields, cross-sectional profiles, and the spatial distribution of absolute errors. At diffusion-dominated regime with $\nu = 0.1$, the second-order diffusion term dominates the energy landscape, resulting in a smooth and widened transition region. The framework achieves its highest precision here, with a relative relative $L^2$ error of only 0.56%. At transition regime with $\nu = 0.1$, the solver continues to provide a robust solution with a relative $L^2$ error of 1.24%, effectively capturing the narrowing of the transition zone. At the most challenging convection-dominated regime with $\nu = 0.01$, where nonlinear convection generates a steep gradient, the framework maintains remarkable stability with a moderate relative $L^2$ error of 4.61%. Crucially, it successfully resolves the sharp discontinuity without the spurious oscillations typically encountered in traditional high-order numerical schemes.

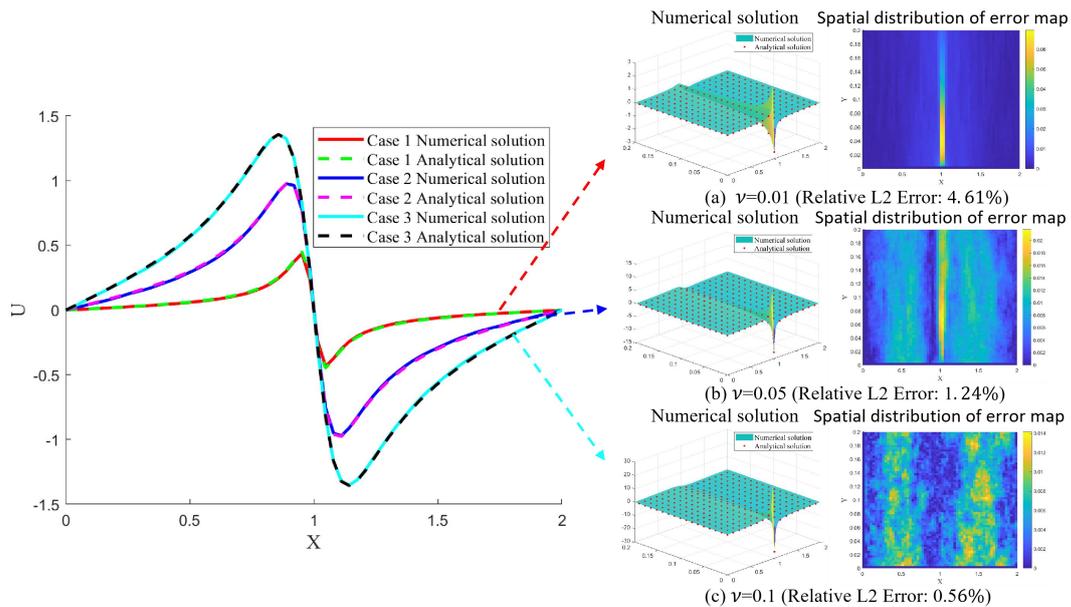

(a) $\nu=0.01$ (Relative L2 Error: 4.61%)

(b) $\nu=0.05$ (Relative L2 Error: 1.24%)

(c) $\nu=0.1$ (Relative L2 Error: 0.56%)

Fig.9 Sensitivity analysis of the viscosity coefficient $\nu$ by comparison of spatiotemporal reconstructions, cross-sectional profiles, and spatial error distributions.

**2) Statistical Reliability and Efficiency Analysis**

To rigorously verify the stability of the proposed method against its stochastic nature, a statistical analysis was performed based on 50 independent trials for each viscosity level. The results, summarized in Table 1, reveal the framework's consistent ability to identify the global minimize from arbitrary high-entropy noise.

The experimental data highlights that the solver consistently converges to the analytical solution with high accuracy. The mean relative $L^2$ errors for $\nu$ = 0.01, 0.05, and 0.1 are 4.572%, 1.068%, and 0.678%, respectively. While the error naturally correlates with the gradient steepness, it remains well within the acceptable range for complex nonlinear reconstruction. Despite starting from purely random initializations, the standard deviations of the errors are extremely low (all below 0.15%). This indicates that the energy-driven iteration is highly deterministic and the final solution is independent of the initial random seed, effectively confirming the global attractor property in nonlinear regimes. Regardless of the nonlinearity intensity or the resulting shock complexity, the average execution time remains consistently under 2 seconds on a 64 × 64 grid. This efficiency underscores the framework's potential for real-time engineering applications where rapid, robust field reconstruction is essential.

Table 1 Statistical performance across 50 independent trials

| Diffusion coefficient $\nu$ | Mean Relative Relative $L^2$ Error(%) | Std. Dev. of Error (%) | Mean Computation Time(s) |
|---|---|---|---|
| $\nu$=0.01 | 4.572 | 0.072 | 1.943 |
| $\nu$=0.05 | 1.068 | 0.121 | 1.394 |
| $\nu$=0.1 | 0.678 | 0.141 | 1.719 |

# 5. Conclusions and future work

This work proposes a randomized PDE energy–driven iterative framework for the efficient and stable solution of partial differential equations. By reformulating governing equations as residual-based energy minimization problems, the method achieves convergence to physically consistent solutions without matrix assembly, training data, or problem-specific tuning. A key observation is that steady elliptic problems exhibit global convergence from arbitrary random initial fields, revealing the strong attractor structure induced by the PDE energy landscape and boundary constraints. For time-dependent diffusion and nonlinear transport equations, the framework employs a space-time unification approach, where the temporal dimension is integrated into a total spatio-temporal energy functional. This formulation naturally recovers the robustness of implicit schemes while enabling the entire solution trajectory to be determined through the same iterative minimization process. Across all cases, Gaussian smoothing and exact boundary enforcement are shown to be essential for stabilizing the iteration and ensuring convergence to the correct physical solution.

The proposed framework provides a flexible foundation for addressing more complex PDE systems. Future work will focus on extensions to multiphysics problems involving coupled fields, non-uniform physical properties, complex and evolving geometries, and general boundary conditions beyond standard Dirichlet and Neumann forms. Further developments will also target large-scale three-dimensional engineering applications, where robustness to initialization, stability under strong nonlinearity, and scalability are critical. These directions position the present approach as a promising alternative paradigm for PDE solution, bridging classical variational methods and modern diffusion-inspired iterative solvers.

**Acknowledgments**

This work was supported by the National Engineering Research Center for High-Speed Railway

**References:**
[1] Evans, L. C. (2010). Partial Differential Equations (2nd ed.). American Mathematical Society.
[2] Brenner, S. C., & Scott, R. (2007). The Mathematical Theory of Finite Element Methods (3rd ed.). Springer Science & Business Media.
[3] Grieves, M., & Vickers, J. (2017). Digital Twin: Mitigating Unpredictable, Undesirable Emergent Behavior in Complex Systems. In Transdisciplinary Perspectives on Complex Systems (pp. 85-113). Springer, Cham.
[4] Tao, F., Zhang, H., Liu, A., & Nee, A. Y. C. (2018). Digital Twin in Industry: State-of-the-Art. IEEE Transactions on Industrial Informatics, 15(4), 2405-2415.
[5] Zienkiewicz, O. C., Taylor, R. L., & Zhu, J. Z. (2015). The Finite Element Method: Its Basis and Fundamentals (7th ed.). Elsevier.
[6] Hughes, T. J. R. (2012). The Finite Element Method: Linear Static and Dynamic Finite Element Analysis. Courier Corporation.
[7] Trottenberg, U., Oosterlee, C. W., & Schüller, A. (2000). Multigrid. Academic Press.
[8] Quarteroni, A., & Valli, A. (1999). Domain Decomposition Methods for Partial Differential Equations. Oxford University Press.
[9] Canuto, C., et al. (2006). Spectral Methods: Fundamentals in Single Domains. Springer.
[10] Gottlieb, S., Ketcheson, D. I., & Shu, C. W. (2009). High Order Strong Stability Preserving Time Stepping Schemes. SIAM Review, 51(3), 509-539.
[11] Benzi, M. (2002). Preconditioning Techniques for Large Linear Systems: A Survey. Journal of Computational Physics, 182(2), 418-477.
[12] Dubey, A., et al. (2014). A Survey of Software Frameworks for Adaptive Mesh Refinement. The International Journal of High Performance Computing Applications, 28(4), 391-409.
[13] Lu, L., Jin, P., Pang, G., Zhang, Z., & Karniadakis, G. E. (2021). Learning nonlinear operators via DeepONet based on the universal approximation theorem of operators. Nature Machine Intelligence, 3(3), 218-229.
[14] Li, Z., et al. (2021). Fourier Neural Operator for Parametric Partial Differential Equations. International Conference on Learning Representations (ICLR)
[15] Li, Z., et al. (2020). Neural Operator: Graph Kernel Network for Partial Differential Equations. Advances in Neural Information Processing Systems (NeurIPS).
[16] Kovachki, N., et al. (2023). Neural Operator: Learning Maps Between Function Spaces. Journal of Machine Learning Research (JMLR).
[17] Raissi, M., Perdikaris, P., & Karniadakis, G. E. (2019). Physics-informed neural networks: A deep learning framework for solving forward and inverse problems involving nonlinear partial differential equations. Journal of Computational Physics, 378, 686-707.
[18] Karniadakis, G. E., et al. (2021). Physics-informed machine learning. Nature Reviews Physics, 3(6), 422-440.
[19] Jagtap, A. D., Kharazmi, E., & Karniadakis, G. E. (2020). Conservative physics-informed neural networks on discrete domains for conservation laws: Advection and Korteweg–de Vries equations. Journal of Computational Physics, 412, 109402.

[20] Jagtap, A. D., Kawaguchi, K., & Karniadakis, G. E. (2020). Adaptive activation functions accelerate convergence in deep and physics-informed neural networks. Journal of Computational Physics, 404, 109136.

[21] Jagtap, A. D., & Karniadakis, G. E. (2020). Extended physics-informed neural networks (XPINNs): A generalized space-time domain decomposition based deep learning framework for nonlinear partial differential equations. Communications in Computational Physics, 28(5), 2002-2041.

[22] Ho, J., Jain, A., & Abbeel, P. (2020). Denoising Diffusion Probabilistic Models. Advances in Neural Information Processing Systems (NeurIPS), 33, 6840-6851.

[23] Song, Y., et al. (2021). Score-Based Generative Modeling through Stochastic Differential Equations. International Conference on Learning Representations (ICLR).

[24] Chung, H., Kim, J., McCann, M. T., Klasky, M. L., & Ye, J. C. (2023). Diffusion Posterior Sampling for General Diffusion Models. International Conference on Learning Representations (ICLR).

[25] Huang, J., et al. (2024). DiffusionPDE: Generative PDE-Solving Under Partial Observation. Advances in Neural Information Processing Systems (NeurIPS).

[26] Psaros, A. F., et al. (2023). Uncertainty quantification in scientific machine learning: Methods, metrics, and comparisons. Computer Methods in Applied Mechanics and Engineering (CMAME), 410, 116002.

[27] Griebel, M., & Knapek, S. (2000). Optimized tensor-product approximation spaces. Constructive Approximation, 16(4), 525-540.

[28] Cannon, J. R. (1984). The One-Dimensional Heat Equation. Cambridge University Press.

[29] Burgers, J. M. (1948). A mathematical model illustrating the theory of turbulence. Advances in Applied Mechanics, 1, 171-199.

[30] Glowinski, R. (1984). Numerical Methods for Nonlinear Variational Problems. Springer-Verlag.

[31] Burman, E. (2007). A continuous finite element method with face penalty for advection-diffusion-reaction problems. Computer Methods in Applied Mechanics and Engineering, 196(21-24), 2337-2362.

[32] Oden, J. T., & Prudhomme, S. (2002). Estimation of modeling error in computational mechanics. Computer Methods in Applied Mechanics and Engineering, 191(23-24), 2637-2647.

[33] Bochev, P. B., & Gunzburger, M. D. (2009). Least-Squares Finite Element Methods. Springer.

[34] R. Temam, Infinite-Dimensional Dynamical Systems in Mechanics and Physics, 2nd ed., Springer, New York, 1997.

[35] J. Hale, Asymptotic Behavior of Dissipative Systems, American Mathematical Society, Providence, RI, 1988.

[36] C. Foias, O. Manley, R. Rosa, and R. Temam, Navier–Stokes Equations and Turbulence, Cambridge University Press, Cambridge, 2001.

[37] G. Roberts and O. Stramer, Langevin-diffusions and Metropolis-Hastings algorithms, Methodology and Computing in Applied Probability, vol. 4, pp. 337–357, 2002.

[38] R. Jordan, D. Kinderlehrer, and F. Otto, The variational formulation of the Fokker–Planck equation, SIAM Journal on Mathematical Analysis, vol. 29, no. 1, pp. 1–17, 1998.

[39] F. Otto, The geometry of dissipative evolution equations: the porous medium equation, Communications in Partial Differential Equations, vol. 26, no. 1–2, pp. 101–174, 2001.

[40] K. Morton, Stability of finite difference approximations to a diffusion-convection equation, International Journal for Numerical Methods in Engineering, vol. 15, no. 5, pp. 677–683, 1980.

[41] C. Canuto, M. Y. Hussaini, A. Quarteroni, and T. A. Zang, Spectral Methods: Evolution to Complex Geometries and Applications to Fluid Dynamics, Springer, Berlin, 2007.

[42] J. von Neumann and R. D. Richtmyer, "A Method for the Numerical Calculation of Hydrodynamic
[43] G. Strang and G. Fix, An Analysis of the Finite Element Method, 2nd ed., Wellesley-Cambridge Press, 2008.